\documentclass[conference]{IEEEtran}
\IEEEoverridecommandlockouts
\usepackage{cite}
\usepackage{amsmath,amssymb,amsfonts}
\usepackage{algorithmic}
\usepackage{graphicx}
\usepackage{textcomp}
\usepackage{xcolor}

 \usepackage{multirow}
\usepackage{amsmath,amssymb,amsfonts}
\usepackage{bm}
\newtheorem{definition}{Definition}
\newtheorem{theorem}{Theorem}
\newtheorem{assumption}{Assumption}
\newtheorem{proposition}{Proposition}

\def\BibTeX{{\rm B\kern-.05em{\sc i\kern-.025em b}\kern-.08em
    T\kern-.1667em\lower.7ex\hbox{E}\kern-.125emX}}
\begin{document}

\title{Distilling Latent Manifolds: Resolution Extrapolation by Variational Autoencoders
}

\author{\IEEEauthorblockN{Jiaming Chu}
\IEEEauthorblockA{\textit{ School of Electronic Engineering} \\
\textit{Beijing University of Posts and Telecommunications}\\
Beijing, China \\
chujiaming886@bupt.edu.cn}
\and
\IEEEauthorblockN{Tao Wang}
\IEEEauthorblockA{\textit{ School of Electronic Engineering} \\
\textit{Beijing University of Posts and Telecommunications}\\
Beijing, China \\
taowang@bupt.edu.cn}
\and
\IEEEauthorblockN{Lei Jin}
\IEEEauthorblockA{\textit{ School of Electronic Engineering} \\
\textit{Beijing University of Posts and Telecommunications}\\
Beijing, China \\
jinlei@bupt.edu.cn}
}

\maketitle

\begin{abstract}
Variational Autoencoder (VAE) encoders play a critical role in modern generative models, yet their computational cost often motivates the use of knowledge distillation or quantification to obtain compact alternatives. Existing studies typically believe that the model work better on the samples closed to their training data distribution than unseen data distribution. In this work, we report a counter-intuitive phenomenon in VAE encoder distillation: a compact encoder distilled only at low resolutions exhibits poor reconstruction performance at its native resolution, but achieves dramatically improved results when evaluated at higher, unseen input resolutions. Despite never being trained beyond $256^2$ resolution, the distilled encoder generalizes effectively to $512^2$ resolution inputs, partially inheriting the teacher model’s resolution preference. We further analyze latent distributions across resolutions and find that higher-resolution inputs produce latent representations more closely aligned with the teacher’s manifold. Through extensive experiments on ImageNet-256, we show that simple resolution remapping—upsampling inputs before encoding and downsampling reconstructions for evaluation—leads to substantial gains across PSNR, MSE, SSIM, LPIPS, and rFID metrics. 

These findings suggest that VAE encoder distillation learns resolution-consistent latent manifolds rather than resolution-specific pixel mappings. This also means that the high training cost on memory, time and high-resolution datasets are not necessary conditions for distilling a VAE with high-resolution image reconstruction capabilities. On low resolution datasets, the distillation model still could learn the detailed knowledge of the teacher model in high-resolution image reconstruction. 
\end{abstract}


\section{Introduction}
Variational Autoencoders (VAEs) are a fundamental component in contemporary generative modeling frameworks, including diffusion~\cite{ho2020denoising,podellsdxl} and flow-based models~\cite{flux}. In such systems, the encoder is responsible for mapping high-dimensional image data into a structured latent space that facilitates efficient generation and manipulation. However, state-of-the-art VAE encoders are often large and computationally expensive, motivating the use of knowledge distillation or quantification to obtain compact alternatives suitable for resource-constrained settings.

While prior work~\cite{sadat2024litevae, yang2025fam, xu2025exploring} on VAE distillation has focused primarily on architectural design and reconstruction fidelity, the evaluation are typically restricted to the corresponding training resolution. The experimental settings of these studies all follow the recognized common sense~\cite{Jansson_2021} in the field: even for model distillation tasks, the knowledge learned by the model is strongly correlated with the sample attributes of the dataset used, and even if the teacher model has stronger generalization ability, the distillation model usually only learns knowledge on the distribution of the training dataset. But in contrast, recent advances in large-scale generative models suggest that VAEs often exhibit strong resolution-dependent characteristics, with optimal performance emerging at particular operating resolutions.

In this paper, we revisit VAE encoder distillation from the perspective of resolution generalization. We conduct a systematic study of a distilled VAE encoder which learn from a large teacher model on only a low-resolution dataset. Surprisingly, we observe that although the distilled encoder performs worse than the teacher when evaluated at its native training resolution, its performance improves dramatically when the input images are upsampled to a higher resolution. This improvement is consistent across multiple reconstruction and perceptual metrics, despite the fact that the student encoder has never been exposed to such resolutions during training.

\begin{figure}[t]
    \centering
    \includegraphics[width=1\linewidth]{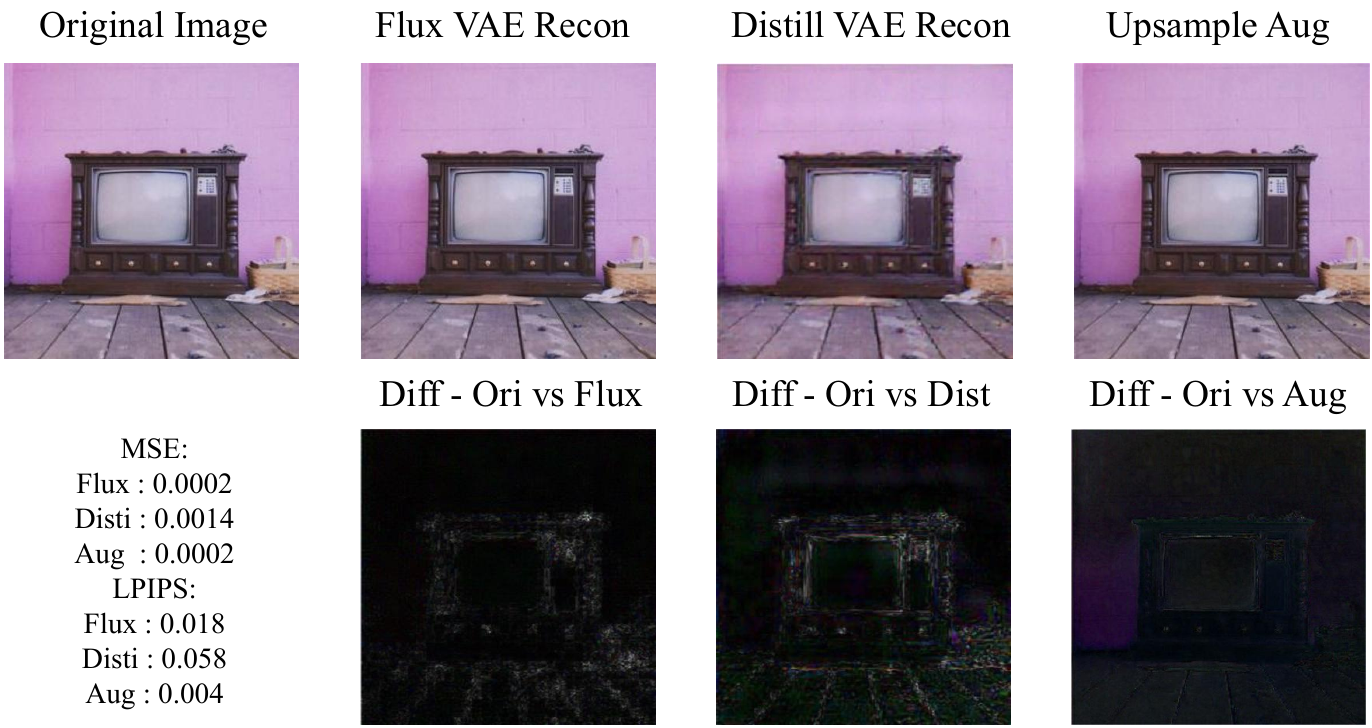}
    \caption{The figure shows low resolution image sample, including the reconstruction results of the Flux~\cite{flux} VAE encoder as the teacher model~(34MB), the distillation model~(2MB) which only training on low resolution images, and the distillation model with the resolution augment. The second row also provides the MSE $\downarrow$, LPIPS distance $\downarrow$ (Lower is better), and differences from the original image for the three.}
    \label{fig:intro}
\end{figure}

To better understand this phenomenon, we analyze the latent representations produced at different input resolutions. We found that even for the same image, the latent variable features obtained by inputting different resolutions into the encoder still have significant differences. This means that the distillation model has the ability to dynamically recognize and extract textures, patterns, and edges under the premise of a certain equivalent receptive field. This observation suggests that the distilled encoder does not merely learn a resolution-specific mapping, but instead captures a resolution-consistent latent manifold inherited from the teacher. As a result, the distilled model partially adopts the teacher’s preferred operating resolution, a behavior we refer to as resolution sweet-spot transfer.

In summary, this work revisits VAE encoder distillation from the perspective of resolution generalization and reveals an overlooked interaction between input resolution and latent representation quality. Rather than learning a resolution-specific mapping, the distilled encoder captures structural properties of the teacher’s latent manifold that manifest more clearly at higher, previously unseen resolutions. Based on these observations, we make the following contributions:

\begin{itemize}
    \item We identify a resolution-induced performance reversal phenomenon in VAE encoder distillation, where a compact encoder distilled at low resolutions exhibits substantially improved reconstruction and perceptual performance when evaluated at higher input resolutions.
    \item We demonstrate that low-resolution distillation enables effective generalization to unseen resolutions, indicating that the distilled encoder learns a resolution-consistent latent representation rather than a pixel-level mapping tied to a specific scale.
    \item We provide empirical evidence that knowledge distillation transfers the teacher’s preferred operating resolution, shedding light on how latent manifold structure and resolution preferences are implicitly inherited by compact generative encoders.
\end{itemize}

We will publicly release the complete project code, model, and training logs on GitHub after the paper is accepted.

\begin{figure*}[th]
    \centering
    \includegraphics[width=0.9\linewidth]{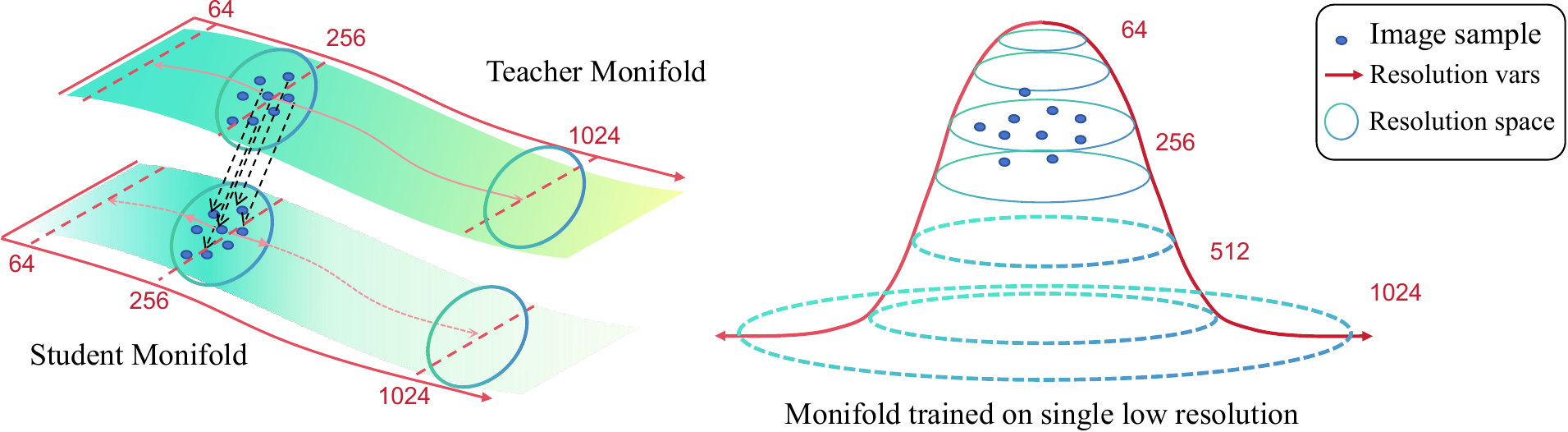}
    \caption{The figure includes the resolution generalization of manifolds during the distillation process of the model and the manifold distribution trained from scratch on low resolution datasets in general. Although distillation and training are carried out at around 256 resolutions, student models can learn the generalization knowledge about high-resolution from teacher models during the distillation process, while models trained from scratch can only achieve generalization around low resolutions. The proof of the distillation model's cross resolution generalization ability and the degree of matching between manifolds can be referred to~\ref{sec:results} and ~\ref{sec:analysis}.}
    \label{fig:manifold}
\end{figure*}

\section{Related works}

\subsection{Design of Lightweight Variational Autoencoder}
Most studies~\cite{mekonnen2024advkdadver,huang2023knowledge,qi2025data,vahdat2020nvae} focus more on the inference cost of step size or denoising modules in compression diffusion or flow processes, and only limit VAE to quantitative methods. Recently, many researchers~\cite{xu2025exploring, Li_2025_WFVAE,wu2025h3ae, liu2023understanding} have been focusing on how to construct lightweight variational autoencoders to replace existing encoder structure designs with smaller and more reasonable constructions. For example, LiteVAE\cite{sadat2024litevae} extracts multi-scale features based on two-dimensional wavelet transform which maintains approximate image reconstruction performance while reducing the number of parameters. The above models are all excellent lightweight VAE solutions, but due to structural differences, their performance is not good when used for distilling existing VAEs.

\subsection{Resolution Generalization and Invariance}
In the field of computer vision, resolution and scale have played an important role in model training, architecture design, and other aspects since the rise of CNN architecture. 
Resolution and scale generalization have been extensively studied in discriminative models, and most researchers~\cite{xu2014scale, xu2020towards,qi2024hierarchical,NIPS2017_38ed162a} adopt multi-scale architectural design or data augmentation to improve robustness to scale variations. In generative modeling, VAEs are commonly interpreted as learning latent manifolds that capture the underlying structure of image distributions~\cite{pmlr-v130-connor21a,cho2023hyperbolic, 10.1162/neco_a_01528}. From this perspective, changes in input resolution can be viewed as alternative parameterizations of the same data manifold. Nevertheless, how distilled VAE encoders generalize to unseen resolutions, and how resolution affects latent alignment with the teacher model, has not been systematically analyzed.

\subsection{Positioning of This Work}
Different from prior distillation studies that focus on model design or fixed-resolution performance, this work investigates the resolution generalization behavior of distilled VAE encoders. By analyzing reconstruction and perceptual metrics across input resolutions, we provide empirical evidence that knowledge distillation can induce resolution-consistent latent representations, revealing an overlooked interaction between distillation and input resolution.

\section{Methodology}

\subsection{Teacher and Student Encoders}
We consider the distillation of a compact VAE encoder from a large teacher model. As the teacher, we adopt the encoder of the Flux VAE, which is trained on large-scale, multi-resolution image data and widely used in modern text-to-image generative models. The teacher encoder outputs a 32-channel latent representation, where the first 16 channels correspond to the mean and the remaining 16 channels correspond to the standard deviation of a Gaussian distribution. Latent codes are obtained via reparameterized sampling.

The student model is designed as a lightweight convolutional encoder that directly predicts a 16-channel latent representation, matching the dimensionality of the sampled teacher latent. Unlike the teacher, the student does not model latent uncertainty explicitly and instead learns a deterministic mapping to the teacher’s latent samples. This design choice allows us to focus on representational alignment while significantly reducing model complexity.

\subsection{Student Architecture}
The student encoder follows a hierarchical residual architecture with progressive downsampling. Given an input image with $C$ channels, the model first applies a $1 \times 1$
convolution to project the input into a low-dimensional feature space with 32 channels. This is followed by a sequence of downsampling stages, each consisting of multiple residual blocks and a strided convolution.

Specifically, the architecture is composed of three downsampling stages. Each stage contains four residual blocks with $3 \times 3$ convolutions, normalization, and identity skip connections, followed by a $3 \times 3$ convolution with stride 2 to reduce spatial resolution while doubling the channel dimension. After the final downsampling stage, a $1 \times 1$ convolution maps the features to a 16-channel latent output. This structure enables the student encoder to progressively aggregate spatial information while maintaining a compact parameter footprint.

Overall, the student encoder contains approximately 2 MB of parameters, compared to approximately 34 MB for the Flux VAE encoder, resulting in a substantial reduction in model size.

\subsection{Distillation Objective}
To align the student’s latent representation with that of the teacher, we employ a direct regression loss between the student output and the sampled teacher latent. Let $z_t$ denote the teacher latent obtained via sampling and $z_s$ denote the student output. The distillation objective is defined using the Huber loss:
\begin{equation}
    L_{distill} = Huber(z_t, z_s;\beta)
\end{equation}

where $\beta$=0.15 controls the transition point between L1 and L2 penalties. Compared to pure L2 loss, the Huber loss provides robustness to outliers in the latent space while preserving sensitivity to small deviations, which we find beneficial for stabilizing distillation.

Notably, no explicit resolution-dependent regularization is introduced during training. The student is supervised only through latent alignment with the teacher at the training resolutions.

\section{Experimental Setup}
\subsection{Dataset}
All experiments are conducted on the ImageNet-256 dataset. For training, we use the standard training split, while evaluation is performed on 50,000 images randomly selected from the validation set. All reported metrics are computed on this fixed evaluation set to ensure consistency across experiments.

\subsection{Implementation details}
Training is performed in two stages to gradually increase the input resolution. In the first stage, input images are downsampled to a resolution of $128^2$. The student encoder is trained for 10,000 steps using a batch size of 64 and the Adam optimizer with a learning rate of $7.5\times10^{-4}$. This stage serves as a warm-up phase that stabilizes training at a lower computational cost.

In the second stage, the input resolution is increased to $256^2$, while all other training settings remain unchanged. The model is trained for an additional 10,000 steps with the same batch size and learning rate. Importantly, the student encoder is never exposed to resolutions higher than $256^2$ during training.

All experiments are conducted on a single NVIDIA RTX 4090 GPU. The first training stage requires approximately 20 minutes with a peak memory usage of 3.3 GB, while the second stage requires approximately 80 minutes with a peak memory usage of 12.3 GB.

We evaluate the distilled encoder using multiple reconstruction and perceptual metrics, including PSNR, MSE, SSIM, LPIPS, and rFID. Reconstruction quality is measured by passing the encoded latent through the corresponding VAE decoder and comparing the output with the ground-truth image.

In addition to standard fixed-resolution evaluation, we consider a resolution-remapped evaluation protocol. In this setting, input images are first upsampled to a higher resolution before encoding, and the reconstructed outputs are downsampled back to the target resolution for metric computation. This protocol allows us to isolate the effect of input resolution on the behavior of the distilled encoder without modifying model parameters or training procedures.

\begin{figure}[t]
    \centering
    \includegraphics[width=1\linewidth]{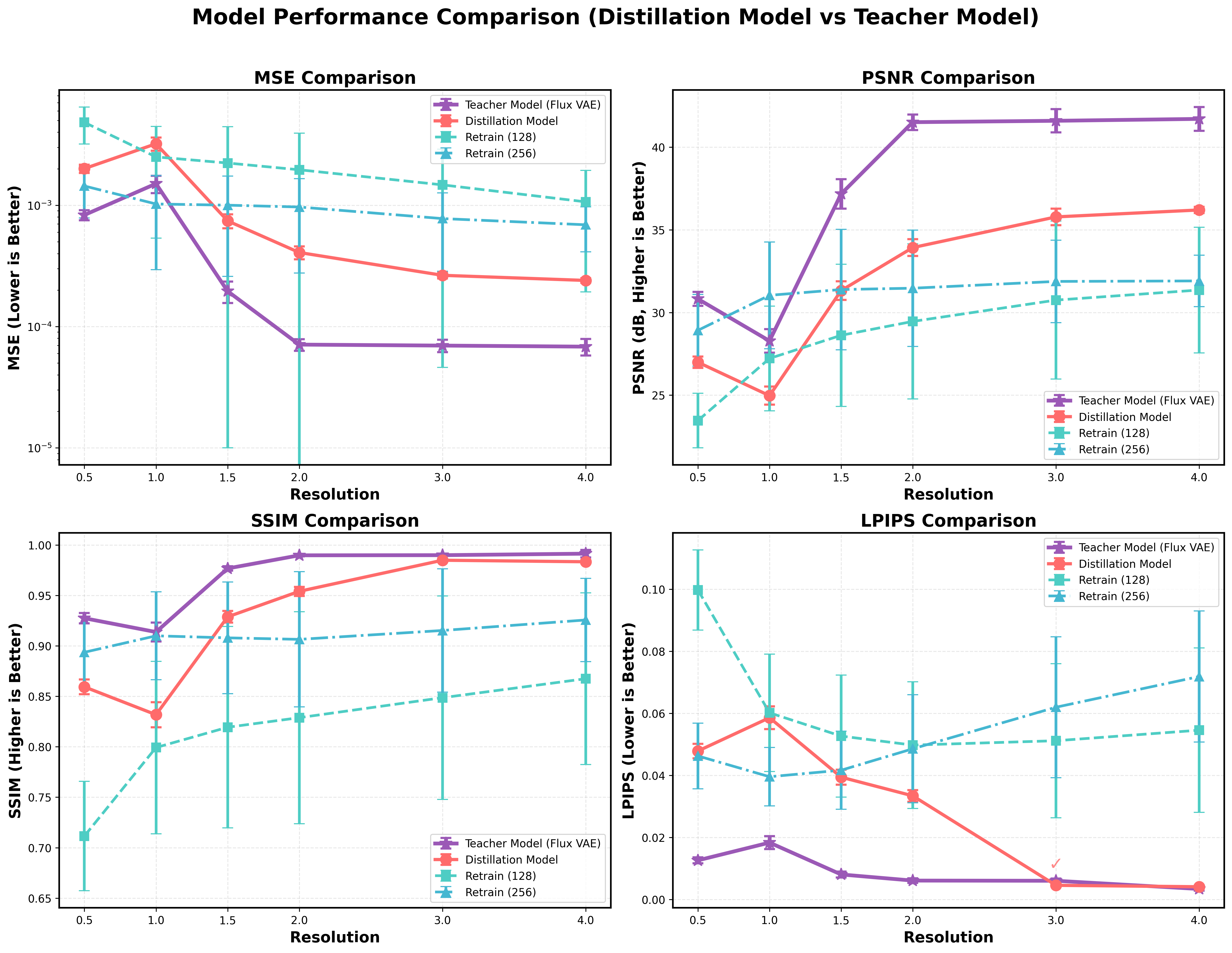}
    \caption{From the figure, we can intuitively observe the consistency between the distillation model~(2MB) and the teacher model~(34MB) in terms of cross resolution performance trends, and the variational encoder~(24MB) trained on low resolution images lacks the ability to generalize across resolutions.}
    \label{fig:Results2}
\end{figure}

\begin{figure}[t]
    \centering
    \includegraphics[width=1\linewidth]{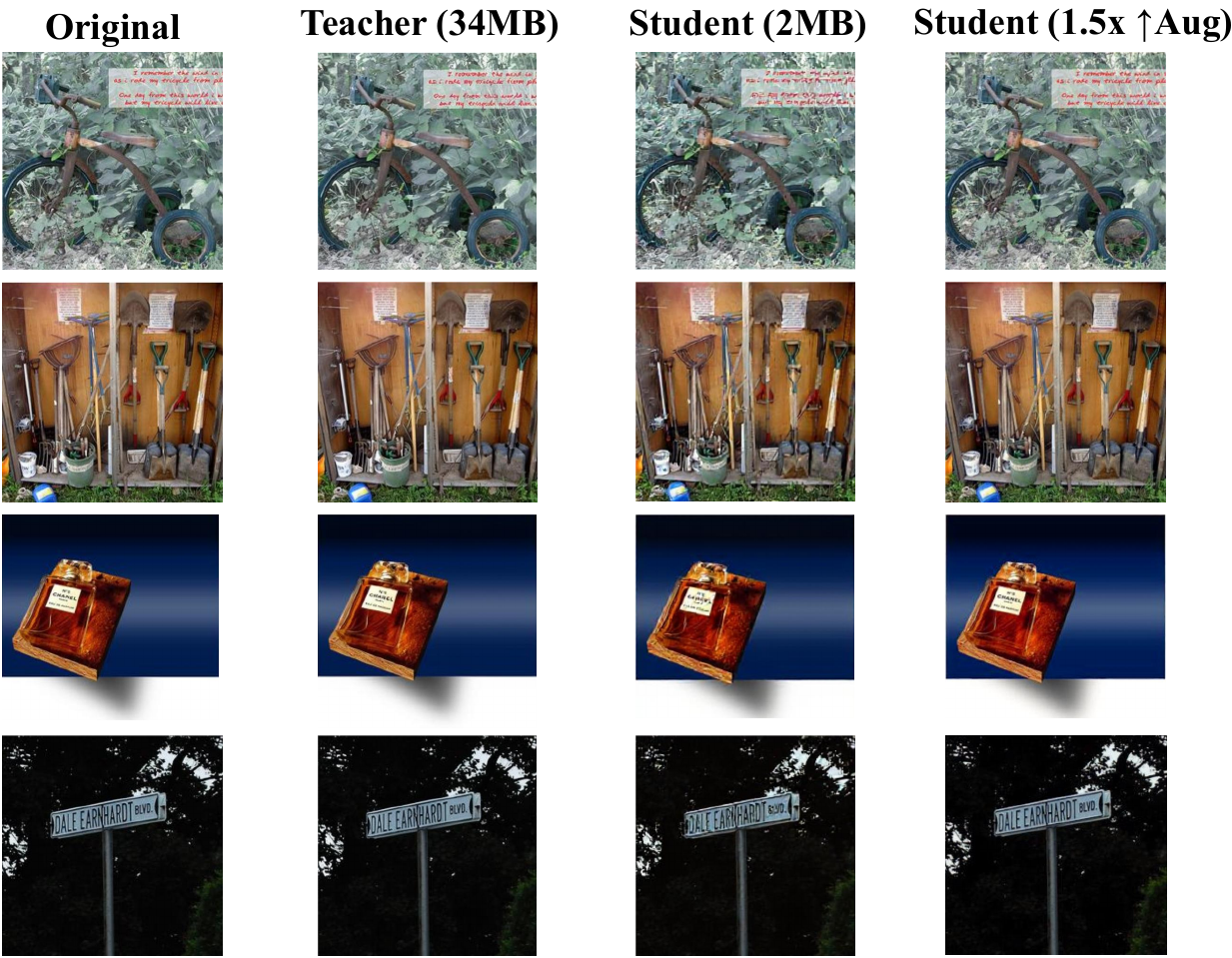}
    \caption{The above figures show the input original image, the reconstruction results of the teacher model, the reconstruction results of the student model, and the image reconstruction results after 1.5-fold upsampling enhancement. From the visualization results, we can see that the student model has little difference from the teacher model in terms of layout, color, brightness, etc., but lacks high-frequency detail information. After upsampling and reinforcement, the reconstruction results surpass the teacher model even in terms of details.}
    \label{fig:vis_result}
\end{figure}

\section{Resolution Generalization Phenomenon}
\label{sec:results}
To systematically investigate the resolution generalization behavior of the distilled encoder, we evaluate both the student and the teacher models under cross-resolution testing. Specifically, input images are upsampled by a scale factor $s \in \{0.5,1,1.5,2,3,4\}$ before encoding, and reconstructed outputs are downsampled back to the original resolution for metric computation. This strategy allows us to isolate the effect of input resolution without modifying model parameters or training procedures.

\begin{table*}[t]
\centering
\caption{Cross resolution test results of distillation model}
\setlength{\tabcolsep}{5mm}
\label{tab:cross_res_results}
\begin{tabular}{ccccccc}
\hline
\textbf{Scale Factor} & \textbf{Model} & \textbf{MSE (×10$^{-4}$)$\downarrow$} & \textbf{PSNR (dB)$\uparrow$} & \textbf{SSIM$\uparrow$} & \textbf{LPIPS$\downarrow$} & \textbf{rFID$\uparrow$} \\
\hline
0.5 & Flux VAE & $8.31_{\pm 0.81}$ & $30.82_{\pm 0.42}$ & $0.927_{\pm 0.005}$ & $0.0126_{\pm 0.0009}$ & 0.069 \\
0.5 & Distilled VAE & $20.04_{\pm 1.57}$ & $26.99_{\pm 0.34}$ & $0.859_{\pm 0.007}$ & $0.0478_{\pm 0.0023}$ & 1.031\\
\hline
1.0 & Flux VAE & $15.05_{\pm 2.45}$ & $28.28_{\pm 0.71}$ & $0.914_{\pm 0.009}$ & $0.0184_{\pm 0.0020}$ & 0.112\\
1.0 & Distilled VAE & $32.11_{\pm 4.04}$ & $24.97_{\pm 0.55}$ & $0.832_{\pm 0.012}$ & $0.0586_{\pm 0.0037}$ & 1.400\\
\hline
1.5 & Flux VAE & $1.96_{\pm 0.39}$ & $37.18_{\pm 0.89}$ & $0.977_{\pm 0.002}$ & $0.0081_{\pm 0.0009}$ & 0.033\\
1.5 & Distilled VAE & $7.43_{\pm 0.97}$ & $31.33_{\pm 0.56}$ & $0.929_{\pm 0.006}$ & $0.0395_{\pm 0.0024}$ & 0.822\\
\hline
2.0 & Flux VAE & $0.71_{\pm 0.08}$ & $41.51_{\pm 0.47}$ & $0.990_{\pm 0.001}$ & $0.0061_{\pm 0.0008}$ & 0.012\\
2.0 & Distilled VAE & $4.08_{\pm 0.49}$ & $33.93_{\pm 0.51}$ & $0.954_{\pm 0.004}$ & $0.0334_{\pm 0.0019}$ & 0.563\\
\hline
3.0 & Flux VAE & $0.70_{\pm 0.08}$ & $41.60_{\pm 0.71}$ & $0.990_{\pm 0.001}$ & $0.0061_{\pm 0.0007}$ & 0.012\\
3.0 & Distilled VAE & $2.64_{\pm 0.15}$ & $35.78_{\pm 0.50}$ & $0.985_{\pm 0.002}$ & $0.0046_{\pm 0.0003}$ & 0.268\\
\hline
4.0 & Flux VAE & $0.68_{\pm 0.11}$ & $41.71_{\pm 0.72}$ & $0.991_{\pm 0.004}$ & $0.0034_{\pm 0.0006}$ & 0.011\\
4.0 & Distilled VAE & $2.40_{\pm 0.10}$ & $36.20_{\pm 0.19}$ & $0.983_{\pm 0.002}$ & $0.0041_{\pm 0.0004}$ & 0.115\\
\hline
\end{tabular}
\end{table*}

\subsection{Cross-resolution behavior of distilled and teacher encoders}
Table~\ref{tab:cross_res_results} summarizes the reconstruction and perceptual performance of the distilled student encoder and the Flux VAE encoder across different input scales. At the native resolution $(s=1\times)$, the student model performs notably worse than the teacher, achieving a PSNR of 24.97~dB compared to 28.28~dB for the Flux VAE, which is expected given the large capacity gap between the two models.

From Fig.~\ref{fig:Results2}, we could find that while the teacher model (Flux VAE, 34 MB) achieves consistent high performance across all scales, the distilled student model (2 MB, trained only on $256^2$ images) shows remarkable improvement from $1\times$ (24.97 dB) to $4\times$ (36.20 dB) scaling. This reveals that the distilled model learns resolution-invariant latent representations that can be more effectively activated by higher-resolution inputs, enabling performance comparable to or exceeding the teacher at original resolution $(s=1\times)$ when tested at $1.5\times$ scale.

Notably, although the student encoder is never trained on resolutions higher than $256^2$, it generalizes effectively to unseen higher-resolution inputs. The improvement is consistent across all reported metrics, suggesting that the observed gains cannot be attributed solely to pixel-level smoothing effects introduced by interpolation. Instead, the results indicate that higher-resolution inputs better align with the latent representation learned through distillation. According to Flux's technical report, Flux VAE has been trained on multiple resolutions and excels at processing images with a resolution of $1024^2$. From Fig.~\ref{fig:Results2}, we can also see that the teacher model shows performance loss at $256^2(1\times)$ resolution, indicating that the teacher model is not good at working at this resolution. However, our distillation model has almost completely learned this characteristic trend, which further indicates that the distilled encoder inherits the resolution-dependent characteristics of the teacher’s latent representation, rather than learning scale-specific features tied to the training resolution.

\subsection{Cross resolution test results of non distillation model}

Although previous studies have shown that models trained at a single resolution do not have scale generalization, in order to ensure that this cross resolution generalization ability does not come from the scale invariance of convolutional structures, we additionally trained two complete variational encoders on the ImageNet-256 dataset for image reconstruction tasks. The two models were trained on image data at 128 and 256 resolutions, respectively, and their hidden channels are up to 128, composing 24MB encoder and 24MB decoder. By observing the performance generalization of these two models at different resolutions, we can determine whether the resolution generalization ability of the distillation model comes from the model architecture design or the generalization knowledge of the teacher model.

From Fig~.\ref{fig:Results2}, we could find the two retrained autoencoders have similar performance curve. From the perspective of changing trends, whatever the training resolution is $128^2$ or $256^2$, as the resolution rise higher, the metrics including MSE, PSNR, SSIM are all be better. Only the metric LPIPS distance goes worse when the scales go over $1.5\times(384^2)$. From the specific performance values, the performance indicators MSE, PSNR, and SSIM show very limited improvement beyond $256^2 (1\times)$ resolution, with almost no improvement. And the LPIPS distance indicator even begins to deteriorate after exceeding 256 resolutions.

These results suggest that the performance gains observed in non-distilled autoencoders primarily stem from increased pixel-space redundancy rather than improved latent representations. The models remain strongly biased toward their training resolution and do not exhibit meaningful generalization to substantially higher resolutions.

\subsection{Comparative analysis and implications}
A clear contrast emerges when comparing distilled and non-distilled models. While all models are trained only at low resolutions, only the distilled encoder exhibits a sustained and monotonic performance improvement that extends well beyond the training resolution range. Moreover, the cross-resolution performance curve of the distilled encoder closely mirrors that of the teacher model, whereas the non-distilled autoencoders show limited and saturating improvements that are tightly coupled to their training resolution.

Additionally, the distilled encoder demonstrates significantly lower variance across all evaluation metrics compared to non-distilled autoencoders, indicating more stable and consistent latent representations across scales. These observations collectively suggest that resolution generalization does not arise from convolutional architecture design alone. Instead, it is induced by latent-level distillation, which aligns the student encoder with a resolution-stable latent manifold learned by the teacher model.

\begin{figure}[t]
    \centering
    \includegraphics[width=1\linewidth]{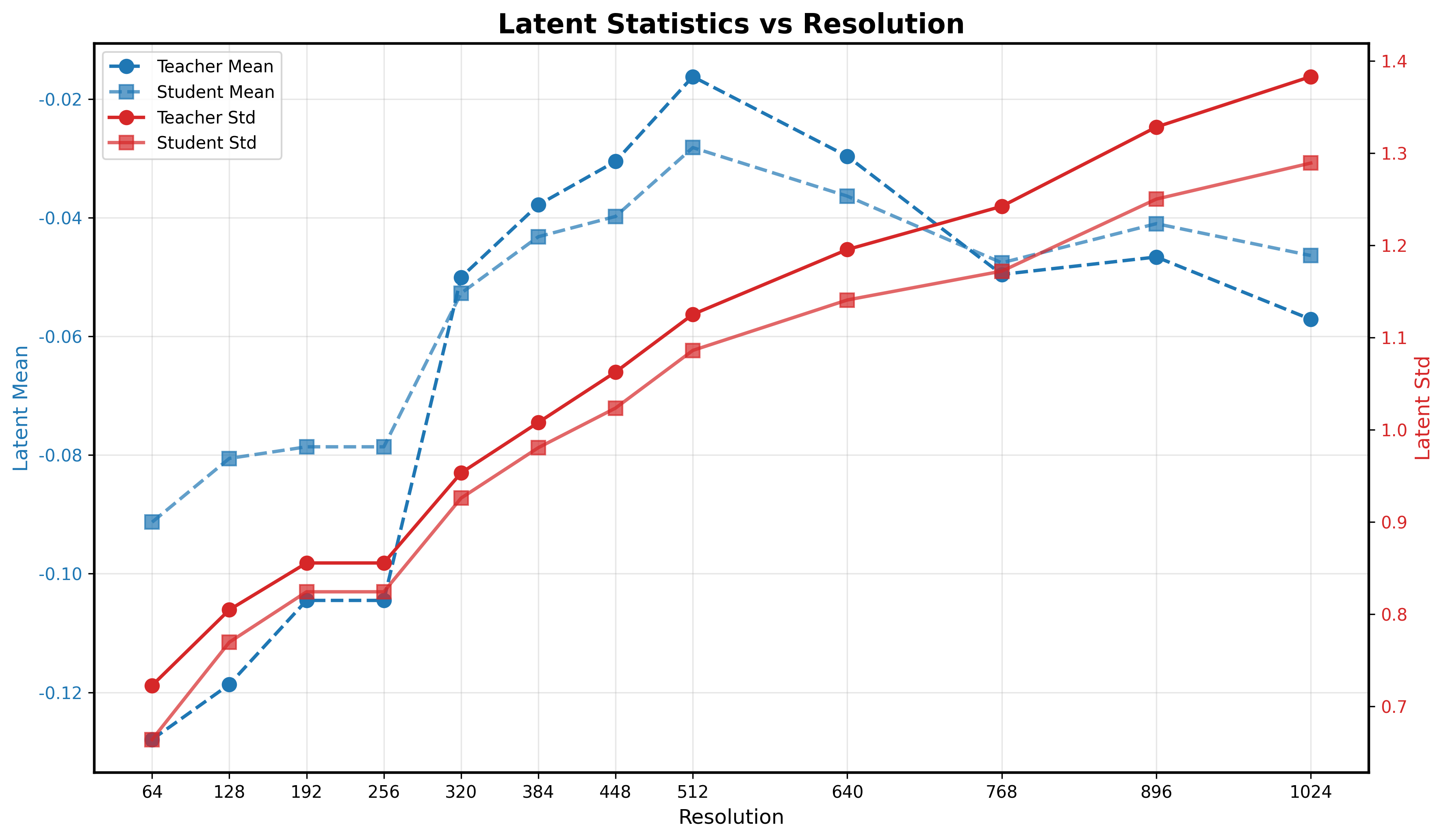}
    \caption{Resolution-dependent latent statistics of the teacher (Flux VAE) and the distilled student encoder. The empirical latent mean and standard deviation are computed over ImageNet validation images at resolutions ranging from $64^2$ to $1024^2$. Both models exhibit highly consistent trends. Despite being trained only on low-resolution images, the student closely follows the teacher’s scaling behavior, indicating that distillation transfers resolution-aware latent parameterization rather than fixed-resolution representations.}
    \label{fig:Latent_Statistics}
\end{figure}

\section{Analysis and Discussion}
\label{sec:analysis}

In this section, we provide a theoretical and empirical analysis to explain the observed resolution generalization phenomenon of the distilled encoder. 

\subsection{Resolution-Dependent Latent Statistics Alignment}
\label{sec:latent_stats}

We first analyze how latent statistics change with resolution. For both the teacher (Flux VAE) and the distilled student encoder, we compute the mean and standard deviation of latent codes across resolutions ranging from $64^2$ to $1024^2$.

As shown in Fig.~\ref{fig:Latent_Statistics}, both models exhibit highly consistent trends in latent mean and variance. Specifically, as input resolution increases, the latent standard deviation increases monotonically, while the latent mean gradually shifts toward zero and stabilizes at higher resolutions. Importantly, the student model closely follows the teacher’s trajectories across all resolutions, with only minor deviations in magnitude.

This observation indicates that distillation transfers not only point-wise latent responses but also the \emph{resolution-dependent scaling law} of the latent space. Despite being trained exclusively on low-resolution images, the student encoder learns how latent activations should expand as a function of input resolution, suggesting the inheritance of a resolution-aware latent parameterization.

\begin{figure}[t]
    \centering
    \includegraphics[width=1\linewidth]{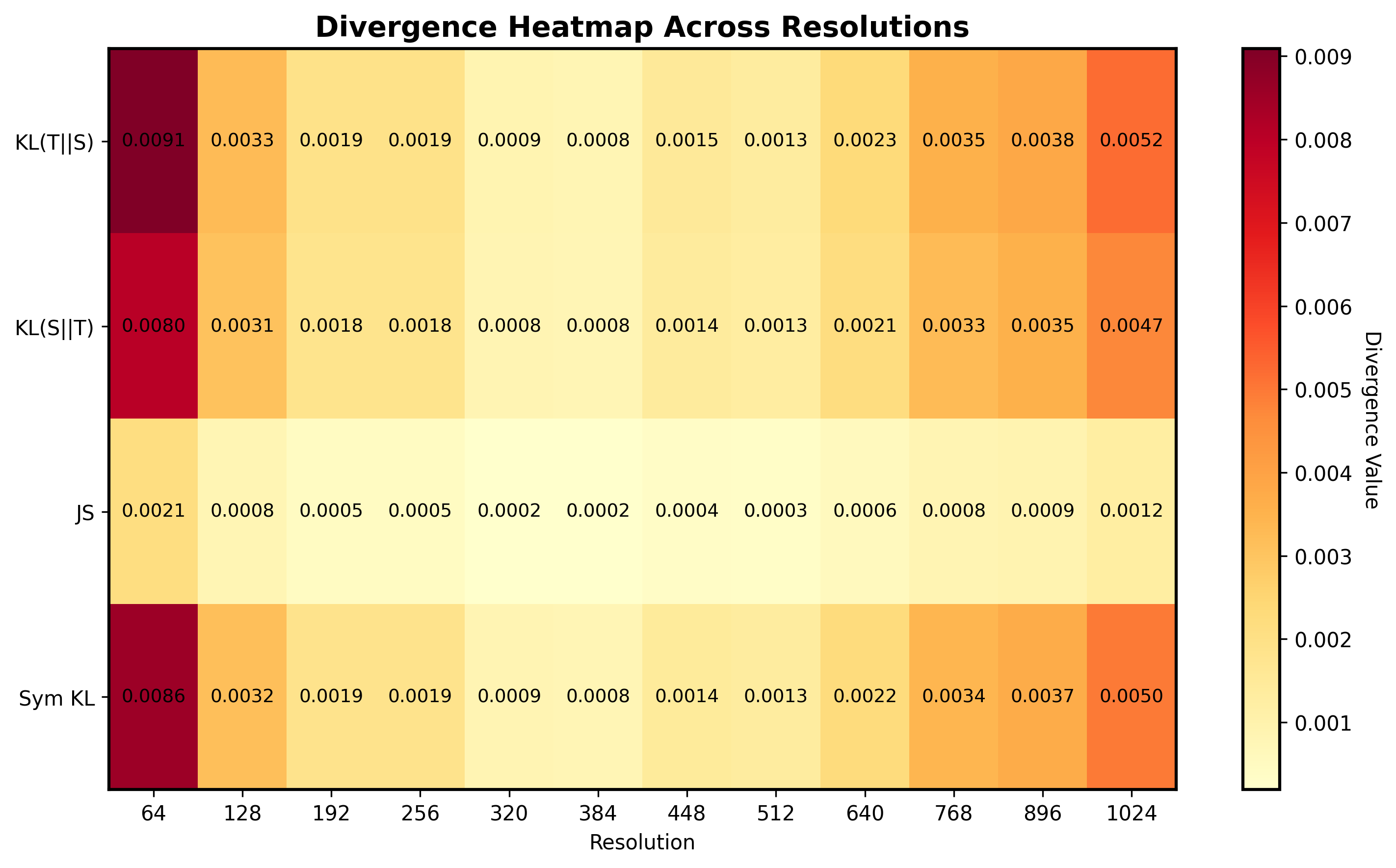}
    \caption{Latent distribution alignment between the teacher and student across resolutions, measured by Kullback--Leibler (KL) and Jensen--Shannon (JS) divergences. Divergence is minimized in the intermediate resolution range ($256^2$–$384^2$), which coincides with the peak reconstruction and perceptual performance of the distilled model. This observation is consistent with the fact that the distillation model is explicitly supervised on 256 resolution data.}
    \label{fig:Divergence}
\end{figure}

\subsection{Latent Distribution Alignment Across Resolutions}
\label{sec:latent_kl}

To quantify the similarity between teacher and student latent distributions, we measure the Kullback–Leibler (KL) divergence and Jensen–Shannon (JS) divergence between their empirical latent distributions at each resolution.

Formally, assuming Gaussian latent distributions $\mathcal{N}(\mu_T, \sigma_T^2)$ and $\mathcal{N}(\mu_S, \sigma_S^2)$, the KL divergence as follow:
\begin{equation}
\mathrm{KL}(T \| S) = \log \frac{\sigma_S}{\sigma_T} + 
\frac{\sigma_T^2 + (\mu_T - \mu_S)^2}{2\sigma_S^2} - \frac{1}{2}.
\end{equation}

As shown in Fig.~\ref{fig:Divergence}, the KL and JS divergences between the teacher and student are minimized in the intermediate resolution range of $256^2$ to $384^2$. Notably, this range coincides with the resolution interval where the distilled model achieves its best reconstruction and perceptual performance.

We refer to this phenomenon as \emph{Resolution Sweet Spot Transfer}: although the student is trained at a single low resolution, distillation aligns its latent manifold such that its optimal operating resolution closely matches that of the teacher. Beyond this sweet spot, divergence increases gradually, primarily due to variance scaling mismatch rather than structural inconsistency.

\begin{figure}[th]
    \centering
    \includegraphics[width=1\linewidth]{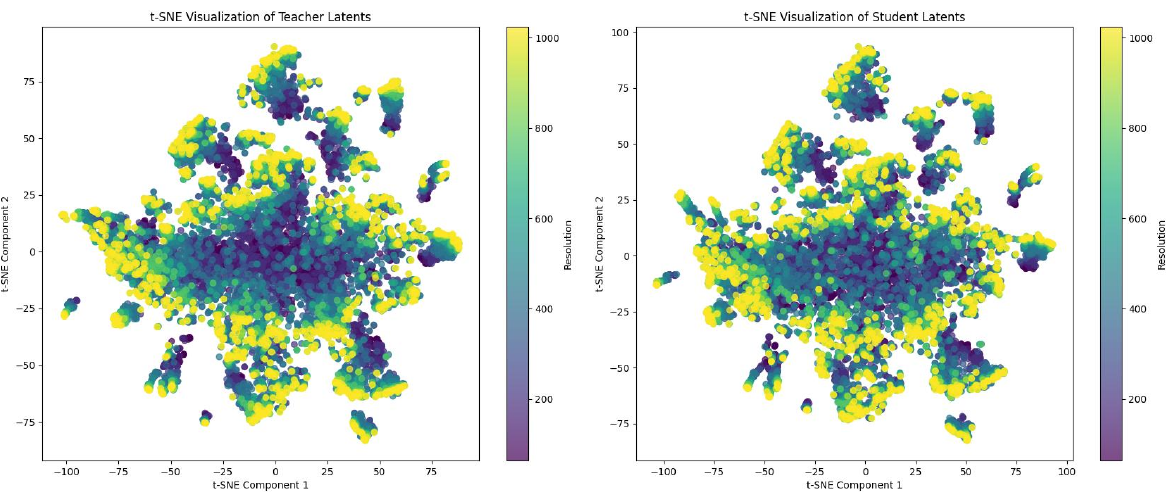}
    \caption{t-SNE visualization of latent embeddings produced by the teacher and distilled student encoders on identical image samples across different resolutions. Latent representations from both models exhibit nearly identical geometric structures, with resolution acting as a radial factor: low-resolution inputs cluster near the center, while higher resolutions progressively expand outward. The strong overlap between teacher and student embeddings demonstrates that distillation successfully transfers the global geometry of the teacher’s latent manifold.}
    \label{fig:tsne}
\end{figure}

\subsection{Geometric Structure of the Latent Manifold}
\label{sec:latent_geometry}

To further examine the geometric organization of latent representations, we visualize the teacher and student latent embeddings using t-SNE, computed on identical image samples across different resolutions.

As illustrated in Fig.~\ref{fig:tsne}, the teacher and student embeddings exhibit remarkably similar global structures. Latent samples corresponding to different resolutions are organized radially: low-resolution inputs concentrate near the center, while higher resolutions progressively expand outward. The highest resolution ($1024^2$) occupies the outermost region of the embedding space.

This visualization provides strong geometric evidence that resolution scaling corresponds to a continuous traversal along a shared latent manifold, rather than transitions between disjoint latent regions. Crucially, the student model reproduces this geometric structure almost identically to the teacher, despite never being exposed to high-resolution training data.

\begin{figure}[th]
    \centering
    \includegraphics[width=1\linewidth]{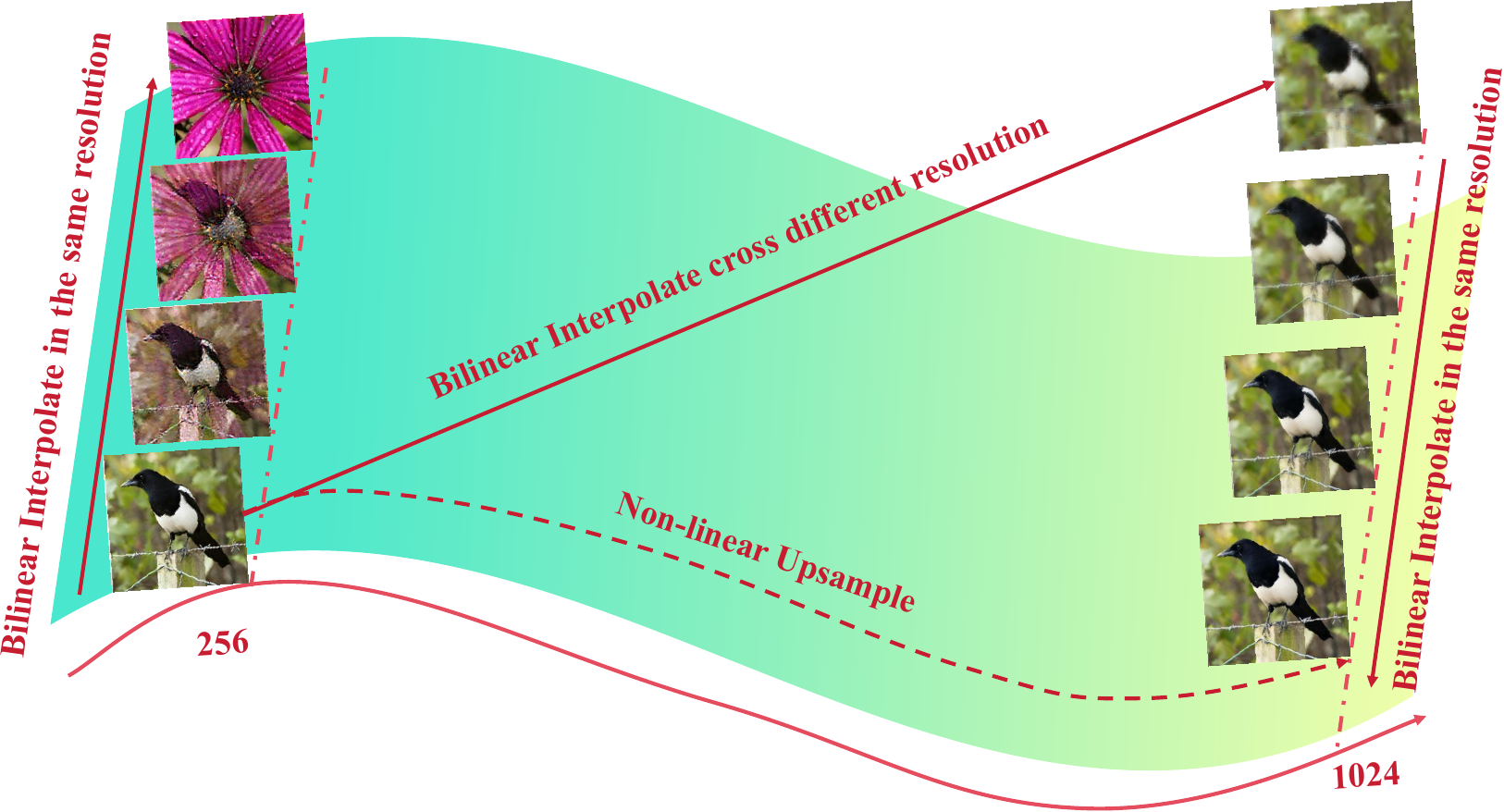}
    \caption{In this figure, we show the results of latent codes interpolation and cross resolution latent codes interpolation in the same resolution direction on the manifold of the student model. From the visualization of interpolation results, we can see that the results of linear interpolation at the same resolution are smooth, while cross resolution interpolation introduces interpolation artifacts.}
    \label{fig:tsne}
\end{figure}

\subsection{Latent Space Continuity Analysis}

To verify that the distilled encoder maintains a smooth latent manifold, we perform linear interpolation between latent codes $z_1 = E_S(x_{r_1})$ and $z_2 = E_S(x_{r_2})$ of the same image at different resolutions:

\begin{equation}
z_{\alpha} = (1-\alpha)z_1 + \alpha z_2, \quad \alpha \in [0,1].
\end{equation}

The interpolated latents are decoded and evaluated for perceptual consistency. Our results show that the distilled encoder maintains smooth transitions, confirming manifold continuity across resolutions.

\section{Theoretical Framework: A Manifold Perspective}
\label{sec:theory}

In this section, we formalize the observed resolution generalization phenomenon through the lens of manifold learning. We provide rigorous mathematical definitions, derive key theoretical results, and establish connections to our empirical findings.

\subsection{Formalization of the Resolution-Manifold Connection}

\begin{definition}[Image Manifold]
Let $\mathcal{M} \subset \mathbb{R}^D$ be a low-dimensional smooth manifold representing the intrinsic structure of natural images, where $D = H_{\max} \times W_{\max} \times C$ is the dimensionality of the highest resolution considered.
\end{definition}

\begin{definition}[Resolution Parameterization]
For each resolution $r = (H_r, W_r) \in \mathcal{R}$, there exists a smooth embedding $\phi_r: \mathcal{M} \to \mathbb{R}^{H_r \times W_r \times C}$ that maps intrinsic coordinates $z \in \mathcal{M}$ to pixel-space images at resolution $r$:
\[
x_r = \phi_r(z)
\]
\end{definition}

\begin{assumption}[Resolution Consistency]
For any two resolutions $r_1, r_2 \in \mathcal{R}$ and intrinsic coordinate $z \in \mathcal{M}$, the images $\phi_{r_1}(z)$ and $\phi_{r_2}(z)$ represent the same semantic content at different scales.
\end{assumption}

\begin{definition}[Encoder as Manifold Learning]
The teacher encoder $E_T: \mathbb{R}^{H \times W \times C} \to \mathcal{Z} \subset \mathbb{R}^d$ learns an approximate inverse of $\phi_r$ composed with a mapping to latent space:
\[
E_T(x_r) = \psi_T(\phi_r^{-1}(x_r)) + \epsilon_T(r) = \psi_T(z) + \epsilon_T(r)
\]
where $\psi_T: \mathcal{M} \to \mathcal{Z}$ is smooth, and $\epsilon_T(r)$ captures resolution-dependent biases.
\end{definition}

\subsection{Geometry of Distillation and Generalization}

\begin{theorem}[Local Tangent Space Alignment]
\label{thm:tangent_alignment}
Assume the teacher's mapping $\psi_T$ is a local diffeomorphism at $z_0 \in \mathcal{M}$. Let $x_{r_0} = \phi_{r_0}(z_0)$ and $E_S^*$ be the optimal student encoder minimizing the distillation loss at resolution $r_0$. Then at $x_{r_0}$:
\[
J_{E_S^*}(x_{r_0}) \approx J_{E_T}(x_{r_0}) \cdot J_{\phi_{r_0}^{-1}}(x_{r_0})
\]
where $J_f$ denotes the Jacobian matrix of function $f$.
\end{theorem}

The distillation objective at resolution $r_0$ is:
\[
\min_{E_S} \mathbb{E}_{z \sim p(z)} \left[ d\left( E_S(\phi_{r_0}(z)), \psi_T(z) + \epsilon_T(r_0) \right) \right]
\]
Let $E_S^*$ be an optimal solution. Taking the functional derivative with respect to $E_S$ and applying the chain rule yields the alignment condition. The approximation arises from neglecting higher-order terms and the bias $\epsilon_T(r_0)$.

\begin{definition}[Resolution Extrapolation Operator]
Define $\Upsilon_{r_0 \to r}: \mathbb{R}^{H_{r_0} \times W_{r_0} \times C} \to \mathbb{R}^{H_r \times W_r \times C}$ as an upsampling operator (e.g., bilinear interpolation).
\end{definition}

\begin{assumption}[Manifold Preservation]
The extrapolation operator approximately preserves manifold structure:
\[
\Upsilon_{r_0 \to r}(\phi_{r_0}(z)) = \phi_r(z) + \eta(r_0, r, z)
\]
where $\|\eta(r_0, r, z)\| \leq \kappa \cdot \|r - r_0\|$ for some $\kappa > 0$.
\end{assumption}

\begin{theorem}[Generalization Error Bound]
\label{thm:error_bound}
For $r > r_0$, the generalization error of the distilled student encoder is bounded by:
\begin{equation*}
\begin{split}
\mathbb{E}_z \| E_S(\Upsilon_{r_0 \to r}(\phi_{r_0}(z))) - E_T(\phi_r(z)) \| 
\leq & \alpha \|r - r_0\| \\
& + \beta \kappa \|r - r_0\| \\
& + \gamma \epsilon_{\text{distill}}
\end{split}
\end{equation*}
where $\alpha$ depends on the Lipschitz constant of $E_S$ and curvature of $\mathcal{M}$. $\beta$ depends on the sensitivity of $E_T$ to manifold perturbations. $\gamma$ is a constant scaling the distillation error $\epsilon_{\text{distill}}$.
\end{theorem}

Decompose the error using the triangle inequality:
\begin{align*}
&\| E_S(\tilde{x}_r) - E_T(x_r) \| \\
&\leq \| E_S(\tilde{x}_r) - E_S(x_{r_0}) \| \quad \text{(Term A)} \\
&+ \| E_S(x_{r_0}) - E_T(x_{r_0}) \| \quad \text{(Term B)} \\
&+ \| E_T(x_{r_0}) - E_T(x_r) \| \quad \text{(Term C)}
\end{align*}
where $\tilde{x}_r = \Upsilon_{r_0 \to r}(x_{r_0})$ and $x_r = \phi_r(z)$.

Term A is bounded by the Lipschitz constant of $E_S$ times $\|\tilde{x}_r - x_{r_0}\|$. Term B is the distillation error. Term C is bounded by the Lipschitz constant of $E_T$ times $\|\phi_{r_0}(z) - \phi_r(z)\|$, which relates to manifold curvature.

\subsection{Resolution Sweet Spot Transfer}

\begin{definition}[Resolution Sweet Spot]
For an encoder $E$, define its resolution sweet spot as:
\[
r_{\text{sweet}}(E) = \arg\min_{r \in \mathcal{R}} \mathbb{E}_z \left[ \mathcal{L}\left( \text{Dec}(E(\phi_r(z))), \phi_{r_{\text{target}}}(z) \right) \right]
\]
where $\text{Dec}$ is the fixed decoder and $\mathcal{L}$ is a reconstruction loss.
\end{definition}

\begin{proposition}[Sweet Spot Transfer]
Under the assumptions of Theorem~\ref{thm:tangent_alignment}, the student's sweet spot approximates the teacher's:
\[
| r_{\text{sweet}}(E_S) - r_{\text{sweet}}(E_T) | \leq C \cdot \epsilon_{\text{align}}
\]
where $\epsilon_{\text{align}} = \mathbb{E}_z \|E_S(\phi_{r_0}(z)) - E_T(\phi_{r_0}(z))\|$ and $C$ depends on the smoothness of $\psi_T$.
\end{proposition}


\section{Additional Ablation Study}
In addition, we also conducted ablation experiments on loss function, resolution, and model capacity, as follows:

\begin{table}[h]
\centering
\caption{The influence of model capacity on distillation performance, MSE should times $10^{-4}$.}
\setlength{\tabcolsep}{2mm}
\label{tab:model capacity}
\begin{tabular}{c|c|c|cccc}
\hline
Resolution           & $C_{hidden}$ & Params & MSE   & PSNR  & SSIM  & LPIPS  \\ \hline
\multirow{3}{*}{256} & 16        & 0.1MB  & 78.21 & 21.56 &                         0.754 & 0.1247 \\
                     & 32        & 2.2MB  & 32.11 & 24.97 & 0.832 & 0.0586 \\
                     & 64        & 6.2MB  & 21.42 & 25.73 &    0.844 & 0.0492 \\ \hline
\end{tabular}
\end{table}

\begin{table}[h]
\centering
\caption{The Influence of Linear Interpolation Types and Positions on Model Performance. "Pre" means the images are resized before input; "Post" means the image are resized after reconstruction, which could observe the effect of only interpolation.}
\setlength{\tabcolsep}{2mm}
\label{tab:interpolate}
\begin{tabular}{c|c|ccccc}
\hline
Interpolation Types           & Position  & MSE$\downarrow$   & PSNR$\uparrow$  & SSIM$\uparrow$  & LPIPS$\downarrow$  \\ \hline
-  & -          & 32.11 & 24.97 &                                    0.832 & 0.0586 \\ \hline
Bilinear  & Post          & 24.62 & 26.15 &                                    0.847 & 0.0548 \\
Bilinear  & Pre       & $7.43$ & $31.33$ & $0.929$ & $0.0395$\\
Bicubic  & Post          & 29.58 & 25.37 &                                    0.840 & 0.0567 \\
Bicubic  & Pre          & 8.18 & 31.23 &                                    0.920 & 0.0303 \\
\hline
\end{tabular}
\end{table}

\begin{table}[h]
\centering
\caption{The influence of different loss functions on model distillation training. Huber loss performs best.}
\setlength{\tabcolsep}{1.5mm}
\label{tab:loss}
\begin{tabular}{l|c|cccc}
\hline
Loss & Training Hours & MSE   & PSNR  & SSIM  & LPIPS  \\ \hline
L1      & 2H  & 42.23 & 23.66 & 0.784   & 0.0862 \\
Huber   & 2H  & 32.11 & 24.97 & 0.832 & 0.0586 \\
Huber+LPIPS    & 16H  & 32.31 & 25.01 &    0.824 & 0.0602 \\ 
Huber+LPIPS+Recon  & 20H  & 38.44 & 23.75 &    0.801 & 0.0634 \\
Huber+LPIPS+KL  & 20H  & 67.02 & 20.31 & 0.730 & 0.1413 \\
                     \hline
\end{tabular}
\end{table}

\begin{table}[h]
\centering
\caption{Encoding efficiency of distillation models compared to other lightweight VAEs (Only Encoder). }
\setlength{\tabcolsep}{1.5mm}
\label{tab:time}
\begin{tabular}{l|ccc|c}
\hline
Model & $T_{infer}$(ms) &params(MB) & $Mem_{GPU}(MB)$  & PSNR \\ \hline
LiteVAE      & 5.32  & 6.6 & 603.1 & 25.01\\
TinyVAE    & 3.67  & 2.4 & 522.2 & 24.68\\ 
FluxVAE  & 8.21  & 34.2 & 794.5 & 28.28\\
Ours  & 3.53  & 2.2 & 488.3 & 24.97\\
Ours(1.5x)  & 4.02  & 2.2 & 540.6 & 31.33\\
                     \hline
\end{tabular}
\end{table}

\section{Conclusion}
This work identifies and analyzes a counter-intuitive phenomenon in VAE encoder distillation: compact encoders distilled at low resolutions exhibit superior performance at higher, unseen resolutions. Through extensive experiments and theoretical analysis, we demonstrate that distillation transfers resolution-aware latent manifolds rather than resolution-specific mappings. Our findings challenge the common assumption that high-resolution training is necessary for high-resolution reconstruction, offering significant practical benefits for resource-constrained applications.

\bibliographystyle{IEEEtran}
\bibliography{main}

\end{document}